\documentclass{article} 

\usepackage{amsmath,amsfonts,bm}

\def\eqref#1{equation~\ref{#1}}

\def\1{\bm{1}}

\def\rx{{\textnormal{x}}}

\DeclareMathAlphabet{\mathsfit}{\encodingdefault}{\sfdefault}{m}{sl}
\SetMathAlphabet{\mathsfit}{bold}{\encodingdefault}{\sfdefault}{bx}{n}

\usepackage{hyperref}
\usepackage{url}
\usepackage{amsmath, amssymb}
\usepackage{algorithm}
\usepackage{algpseudocode}
\usepackage{xcolor}
\usepackage{setspace}
\usepackage{iclr2026_conference,times}
\usepackage{tikz}
\usepackage{amsmath}
\usepackage{pgfplots, pgfplotstable}
\usepackage{subcaption}
\usepackage{color,soul}
\usepackage[most]{tcolorbox}
\usepackage{placeins}
\usetikzlibrary{arrows.meta, fit, calc, backgrounds}

\newcommand{\vect}{\mathbf}
\newcommand{\vectorproj}[2][]{\textit{proj}_{\vect{#1}}\vect{#2}}

\pgfplotsset{compat=1.18}
\usetikzlibrary{shapes.geometric, arrows.meta, positioning, angles, quotes}

\tikzstyle{block} = [rectangle, draw, minimum height=1.2em, minimum width=2em, text centered]
\tikzstyle{cloud} = [ellipse, draw, text centered, minimum height=2em]
\tikzstyle{arrow} = [thick, ->, >=stealth]
\tikzstyle{actor} = [circle, draw, minimum size=1.2em, inner sep=0pt]

\title{What Is Missing: Interpretable Ratings for Large Language Model Outputs}

\author{
Nicholas Stranges \quad Yimin Yang\\
Department of Electrical and Computer Engineering\\
Western University, Canada\\
\texttt{nstrang2@uwo.ca}\qquad\texttt{yimin.yang@uwo.ca}
}

\iclrfinalcopy 
\begin{document}

\maketitle

\begingroup
\renewcommand{\thefootnote}{}%
\footnotetext{Code: \url{https://github.com/nstranges/what-is-missing/tree/main}}%
\addtocounter{footnote}{-1}%
\endgroup

\begin{abstract}
 Current Large Language Model (LLM) preference learning methods such as Proximal Policy Optimization and Direct Preference Optimization learn from direct rankings or numerical ratings of model outputs. These rankings are subjective, and a single numerical rating chosen directly by a judge is a poor proxy for the quality of natural language. We introduce the \emph{What Is Missing} (WIM) rating system to produce rankings from natural-language feedback. WIM integrates into existing training pipelines, can be combined with other rating techniques, and can be used as input to any preference learning method without changing the learning algorithm. To compute a WIM rating, a human or LLM judge writes feedback describing what the model output is missing. We embed the output and the feedback with a sentence embedding model and compute the cosine similarity between the resulting vectors. We empirically observe that, compared to discrete numerical ratings, WIM yields fewer ties and larger rating deltas, which improves the availability of a learning signal in pairwise preference data. We use ``interpretable'' in the following limited sense: for each scalar rating, we can inspect the judge's missing-information text that produced it, enabling qualitative debugging of the preference labels.
\end{abstract}

\section{Introduction}

The creation of the Large Language Model (LLM) has changed what humans can do with a computer \cite{NEURIPS2020_1457c0d6}. To achieve these technological breakthroughs,  a large corpus of data and training resources is required \cite{kandpal2025position}. The training time and resources are split between two distinct phases: pre-training and post-training. An LLM that has been pre-trained is an excellent next word prediction machine and has some ability to perform instruction-following tasks \cite{radford2019language}.

The second phase, post-training, can be broken into two categories: the Supervised Fine-Tuning (SFT) phase and the preference learning phase \cite{fernando2025mitigatingforgettingllmsupervised}. The SFT phase can train an LLM to produce specific outputs by minimizing a cross entropy loss on an instruction-following dataset. The preference learning phase aims to improve the usefulness of the LLMs by tuning the model to human preferences \cite{ouyang2022training}. As human preferences cannot be directly calculated, the preference learning phase requires the use of a reward model and reinforcement learning (RL) instead of a direct loss function \cite{ouyang2022training}.

Expanding the post-training tool set will allow researchers to better prevent misalignment. Misalignment is described as the difference between human goals and the objectives of the LLM \cite{10.5555/3294996.3295184}. Misalignment is an ever-increasing problem as model performance continues to improve and is crucial as the newest models have warranted new safety protections, such as Anthropic’s AI Safety Level 3 (ASL-3) designation for Claude Opus 4 \cite{anthropic_asl3_2025}. If models acquire superhuman intelligence, they must be aligned to human goals and values or there is a potential for catastrophic damage to human civilization \cite{carlsmith2024powerseekingaiexistentialrisk}.

One of the primary tools for addressing misalignment is preference learning, where the training loop revolves around ranking model outputs and optimizing the model on that ranking. Historically, the ranking system was decided using direct rankings as human evaluators would directly rank completions using their own preferences \cite{ouyang2022training}. Rankings are subjective, relying on heuristic evaluations and user preferences rather than clear performance metrics \cite{kumar2025llmposttrainingdeepdive}. A method to understand why every ranking was chosen is impossible as different judges will not always create the same ranking. An improvement on the ranking system is to score each output using a numerical rating system such as a scale from 1-10, as seen in \cite{10.5555/3692070.3693141}. This allows outputs in a ranking to be compared with each other and can demonstrate how much better or worse responses are compared with each other. 

Fundamentally, a numerical rating system has the same shortfalls as a direct ranking system because it is difficult to distill the worth of an answer into a single number. Outputs with the same rating can differ significantly. A numerical rating system is a discrete set and is a poor tool to quantify a complex system such as human language. Experiments in Section~\ref{rating_diffs} empirically demonstrate that numerical ratings can produce the same rating in a pairwise comparison, preventing the generation of a learning signal. As the use of synthetic data increases, other LLM systems have taken the role of the judge \cite{pmlr-v235-lee24t}. LLM feedback systems still rely on the same methods to rank and therefore, post-train models.

Overall, shortcomings of the existing ranking and rating systems can be classified into the following categories: the low interpretability of the ratings and the same ratings preventing the creation of a learning signal. This paper introduces \emph{What Is Missing} (WIM) feedback scoring as an alternative to traditional numerical ratings or direct rankings. WIM provides natural language feedback, making the rating directly interpretable. The produced rating distribution are discrete samples of a continuous distribution and therefore, repeated ratings are much less frequent. Both of these improvements position WIM as a solution to increase preference learning performance while simultaneously providing interpretable ratings.


\section{The Proposed WIM Method}

\tikzset{
  myarrow/.style={
    -{Stealth[length=2mm, width=2mm]},
    line width=1.1pt,
    rounded corners
  },
  ghostarrow/.style={-,
                     line width=1.1pt, rounded corners,
                     draw opacity=0.25}
}

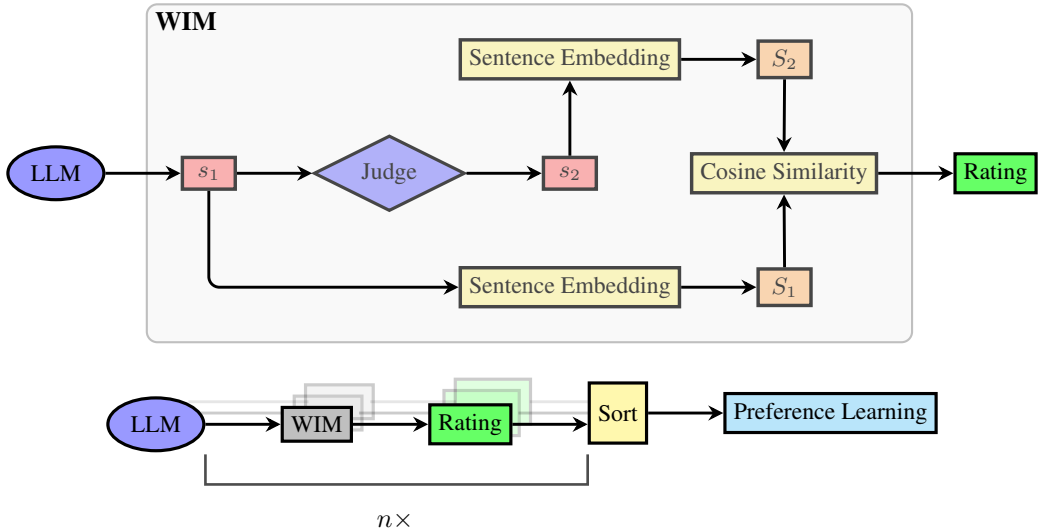
\begin{figure}[ht] 
\begin{center}
\begin{tikzpicture}[font=\small]
  \node (llm) [cloud, fill=blue!40!white, line width=1.2pt] {LLM};
  \node (s1) [block, right=of llm, fill=red!40!white, line width=1.3pt] {$s_1$};
  \node (critic) [diamond, draw, aspect=2, right=of s1, fill=blue!40!white, line width=1.2pt] {Judge};
  \node (s2) [block, right=of critic, fill=red!40!white, line width=1.3pt] {$s_2$};
  \node (embedS2) [block, above=of s2, fill=yellow!40!white, line width=1.3pt] {Sentence Embedding};
  \node (embedS1) [block, below=of s2, fill=yellow!40!white, line width=1.3pt] {Sentence Embedding};
  \node (vecS2) [block, right=of embedS2, fill=orange!40!white, line width=1.3pt] {$S_2$};
  \node (vecS1) [block, right=of embedS1, fill=orange!40!white, line width=1.3pt] {$S_1$};
  \node (cosine) [block, right=2.97cm of critic, yshift=0.0cm, fill=yellow!40!white, line width=1.3pt] {Cosine Similarity};
  \node (wim) [block, right=of cosine, fill=green!60!white, line width=1.2pt] {Rating};

  \node[fill=gray!15, fill opacity=0.3,rounded corners,fit=(s1)(critic)(s2)(embedS2)(embedS1)(vecS2)(vecS1)(cosine),
        inner sep=0.45cm, draw=gray!50, line width=0.8pt] (background) {};
    \node[anchor=north west, font=\bfseries]
       at (background.north west) {WIM};

  \draw [myarrow] (llm) -- (s1);
  \draw [myarrow] (s1) -- (critic);
  \draw [myarrow] (critic) -- (s2);
  \draw [myarrow] (s2) -- (embedS2);
  \draw [myarrow] (embedS2) -- (vecS2);
  \draw [myarrow] (s1) |- (embedS1);
  \draw [myarrow] (embedS1) -- (vecS1);
  \draw [myarrow] (vecS1.north) -- (cosine.south);
  \draw [myarrow] (vecS2.south) -- (cosine.north);
  \draw [myarrow] (cosine.east) -- (wim.west);
\end{tikzpicture}

\bigskip

\begin{tikzpicture}[font=\small]
  \node (llm) [cloud, fill=blue!40!white, line width=1.2pt] {LLM};
  \node (wim) [block, right=of llm, fill=gray!50, line width=1.2pt] {WIM};
  \node (rating) [block, right=of wim, fill=green!60!white, line width=1.2pt] {Rating};
  \node (argmax) [block, right=of rating, fill=yellow!40!white, line width=1.2pt, minimum height=0.8cm, yshift=0.15cm] {Sort};
  \node (winning) [block, right=of argmax, fill=cyan!25!white, line width=1.2pt] {Preference Learning};

  \begin{scope}[on background layer]
    \foreach \i in {1,...,2} {
      \node[block,
            fill=gray!40!white,
            fill opacity=0.2,
            draw opacity=0.2,
            line width=1.2pt,
            minimum width=0.9cm,
            minimum height=0.45cm]
          at ($(wim.center)+(0.15*\i,+0.15*\i)$) {};
      \node[block,
        fill=green!60!white,
        fill opacity=0.2,
        draw opacity=0.2,
        line width=1.2pt,
        minimum width=1.0cm,
        minimum height=0.6cm]
      at ($(rating.center)+(0.15*\i,0.15*\i)$) {};
    }
  \end{scope}

    \begin{scope}[on background layer]
      \foreach \i/\op in {1/.20, 2/.10}{
        \draw[ghostarrow, draw opacity=\op, shorten >=-2pt, shorten <=-15pt]
          ($ (llm.east) + (0.15*\i, 0.15*\i) $) --
          ($ (wim.west) + (0.15*\i, 0.15*\i) $);
        \draw[ghostarrow, draw opacity=\op, shorten >=-2pt, shorten <=-2pt]
          ($ (wim.east) + (0.15*\i, 0.15*\i) $) --
          ($ (rating.west) + (0.15*\i, 0.15*\i) $);
        \draw[ghostarrow, draw opacity=\op, shorten >=-2pt, shorten <=-2pt]
          ($ (rating.east) + (0.15*\i, 0.15*\i) $) --
          ($ (argmax.west)!(rating.east)!(argmax.west) + (0.15*\i, 0.15*\i) $);
      }
    \end{scope}

  \draw [myarrow] (llm) -- (wim);
  \draw [myarrow] (wim) -- (rating);
  \draw [myarrow] (rating.east) -- ($(argmax.west)!(rating.east)!(argmax.west)$);
  \draw [myarrow] (argmax) -- (winning);

  \draw[draw=black!70, line width=1pt]
  ($(llm.east)+(0,-0.4cm)$)   
  -- ++(0,-0.4cm)             
  -- ($(argmax.west)+(0,-0.95cm)$) 
  -- ++(0,0.4cm);             
    \node[font=\bfseries] at ($(rating.south)+(-1cm,-1cm)$) {$n\times$};
\end{tikzpicture}
\end{center}
\caption{Flowchart of the WIM method. An LLM produces a natural language output \(s_1\). \(s_1\) is then evaluated by a human or an LLM judge. The judge's goal is to produce \(s_2\), a response containing what is missing in \(s_1\). Both \(s_1\) and \(s_2\) are passed through a sentence embedding model to produce high dimensional vectors \(S_1\) and \(S_2\). The similarity of \(S_1\) and \(S_2\) is calculated using cosine similarity and the resulting similarity score is the WIM rating. A higher similarity between \(S_1\) and \(S_2\) implies that there is less missing from the LLM's output. $n$ model outputs are rated by the WIM method and then sorted to produce a ranking. The ranking of the outputs is then passed to a preference learning algorithm.}
\label{figure1}
\end{figure}

The process of creating \emph{What Is Missing} (WIM) feedback scoring is demonstrated in Figure~\ref{figure1}. In WIM, a human or LLM judge produces a natural-language description of what was missing in the model's output. For example, if the model forgets to mention a keypoint in its argument or forgot some functionality when performing a coding task.

Conceptually, this process is adversarial: the model aims to include all relevant information, while the judge identifies missing elements.  
This dynamic is similar to the discriminator in a Generative Adversarial Network, although here the goal is to surface missing content rather than distinguish between real and generated examples \cite{NIPS2014_f033ed80}.

The scoring procedure works as follows:  
\begin{enumerate}
    \item The base model output (\(s_1\)) and the WIM response (\(s_2\)) are each passed through a sentence embedding model, producing high-dimensional vector representations \(S_1\) and \(S_2\) \cite{reimers2019sentencebert}. These embeddings capture semantic properties of each text.
    \item Cosine similarity is computed between \(S_1\) and \(S_2\) to quantify semantic overlap \cite{mikolov2013efficientestimationwordrepresentations}.
    \item The resulting score, in \([-1, 1]\), serves as the feedback rating for the base model’s output.  
          If no WIM feedback is provided (i.e., nothing was missing), a perfect score of \(1\) is assigned as a design choice.
    \item Once a WIM score for all outputs being compared has been computed, the scores are ranked from highest to lowest. The ranking can then be used as input to any preference learning algorithm such as Proximal Policy Optimization (PPO) or Direct Preference Optimization (DPO) \cite{schulman2017proximalpolicyoptimizationalgorithms} \cite{rafailov2023direct}.
\end{enumerate}

\subsection{Mathematical Explanation}

Referring again to Figure~\ref{figure1}, let the model’s generated output be a sequence of \(n\) tokens:

\begin{equation}
    s_1 = [w_1, w_2, \ldots, w_n],
\end{equation}

where \(w_i\) is the \(i\)-th token.

The WIM response is a sequence of \(m\) tokens describing what \(s_1\) omitted:

\begin{equation}
    s_2 = [w'_1, w'_2, \ldots, w'_m].
\end{equation}

A sentence embedding function \(f_{\text{embedding}}\) maps each sequence into a vector in \(\mathbb{R}^d\):

\begin{equation}
    S_1 = f_{\text{embedding}}(s_1) \in \mathbb{R}^d, \quad
    S_2 = f_{\text{embedding}}(s_2) \in \mathbb{R}^d.
\end{equation}

The WIM score is computed as the cosine similarity between these vectors:
\begin{equation}
    \text{WIM} = \frac{S_1 \cdot S_2}{\|S_1\| \, \|S_2\|}.
\end{equation}

A higher WIM score indicates that the model’s output and the WIM feedback are more semantically similar, suggesting less \emph{missingness}.

\subsection{Missingness}

The WIM vector, $S_2$ can be decomposed as shown in Equation~\ref{S2Decomposition}:

\begin{equation} \label{S2Decomposition}
S_2 = \underbrace{\vectorproj[S_1]{S_2}}_{\text{parallel feedback}}+\underbrace{(S_2-\vectorproj[S_1]{S_2})}_{\text{orthogonal feedback}} = S_2^{\parallel} + S_2^{\perp}
\end{equation}

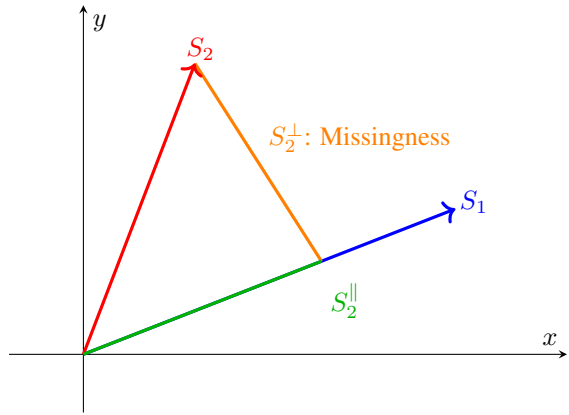
\begin{figure}[ht]
\centering
\begin{tikzpicture}
  \begin{axis}[
      axis lines=middle,
      xlabel={$x$}, ylabel={$y$},
      xmin=-0.2, xmax=1.3,
      ymin=-0.2, ymax=1.2,
      xtick=\empty, ytick=\empty,
      width=9cm, height=7cm
    ]

    \addplot[->, very thick, blue]
      coordinates {(0,0) (1.0,0.5)} node[pos=1.05] {$S_1$};
      
    \addplot[-, very thick, orange]
      coordinates {(0.64,0.32) (0.3,1.0)}
      node[midway, above right] {$S_2^{\perp}$: Missingness};

    \addplot[->, very thick, red]
      coordinates {(0,0) (0.3,1.0)} node[pos=1.05] {$S_2$};

    \addplot[-, very thick, green!70!black]
      coordinates {(0,0) (0.64,0.32)}
      node[pos=1.1, anchor=north, yshift=-8pt] {$S_2^{\parallel}$};
  \end{axis}
\end{tikzpicture}
\caption{2D visualization of missingness}
\label{missingness_visualization}
\end{figure}

\emph{Missingness} represents the missing content in the model output and can be thought of as the \emph{orthogonal feedback vector}, $S_2^{\perp}$. As this vector grows in relation to the \emph{parallel feedback vector}, $S_2^{\parallel}$, the amount of missing content in the model's response should grow proportionally. Orthogonality to $S_1$ implies $S_1^{\top} S_2^{\perp} = 0$ and is equivalent to having no common information in an embedding space. As $S_2^{\perp}$ grows in relation to $S_2^{\parallel}$, the angle between vectors $S_1$ and $S_2$ will also increase as the missingness vector's magnitude is given as $\| S_2^{\perp}\| = \| S_2 \| \sqrt{1 - \cos^{2}\theta}$ (Appendix~\ref{missingness_derive}). The angle in the $\| S_2^{\perp}\|$ means cosine similarity is a valid metric to measure missingness. The relationship between missingness and cosine similarity can be visualized on the 2D plane in Figure~\ref{missingness_visualization}. Note the same interpretation can be taken even when $S_2^{\parallel}$ is antiparallel as that case would result in a negative WIM rating.

\subsection{Ranking Usage}
\label{method_agnostic}

Online Direct Preference Optimization (ODPO) was chosen as the preference learning algorithm to optimize the model for the selected ranking \cite{guo2024directlanguagemodelalignment}. ODPO was chosen because it does not require the training of a reward model and, since it is online, the model can be trained using LLM inference instead of creating a dataset of responses.

As WIM aims to improve how the ranking of model output is determined, ODPO is not required for this process. WIM is agnostic to the training process as it only aims to improve the rankings for any existing preference learning algorithm such as Proximal Policy Optimization (PPO) and Group Relative Policy Optimization (GRPO) \cite{schulman2017proximalpolicyoptimizationalgorithms} \cite{shao2024deepseekmathpushinglimitsmathematical}. Being algorithm agnostic allows WIM to be directly implemented into existing training infrastructure, saving engineering costs and the time to launch. Other natural language feedback systems such as Text2Grad require a completely new training process including training a separate reward model \cite{wang2025text2gradreinforcementlearningnatural}. Alternative training techniques with unique post-training approaches such as Constitutional AI, still contain a preference learning phase after a unique SFT phase \cite{bai2022constitutionalaiharmlessnessai}. As new preference learning methods are created, WIM will continue to be useful as long as preference learning is based on rankings. 

\subsection{Self-Judging}

There is no requirement that a human produces the WIM response. A larger and more powerful model can produce the WIM feedback, and the model being trained can also act as a self-judge to reflect on its own output \cite{10.5555/3692070.3694459}. We view self-judging as analogous to a researcher revising a first draft: the same author generates an output and then critiques what is missing.

In this paper, we reuse the same LLM for acting and judging by switching the context and instructions. We evaluate two self-judge configurations:
(i) \emph{Fixed Judge}: the frozen reference model \(\pi_{\text{ref}}\) generates WIM critiques; and
(ii) \emph{Moving Judge} (a.k.a. ``changing judge''): the current, actively updated model \(\pi_{\theta}\) generates WIM critiques.

We expect these settings to behave differently because a moving judge changes the critique distribution during training: improvements (or regressions) in the actor also change the judge, which can create non-stationary targets and potentially unstable feedback. In contrast, a fixed judge provides a stable critique distribution, which may yield more stable optimization even if the judge is weaker. We treat this explanation as a hypothesis and return to it in the discussion of results and limitations.

\section{Theoretical Analysis}

The WIM feedback system is interpretable in the narrow, data-centric sense that each scalar score is directly derived from an accompanying natural-language ``what is missing'' critique. This makes the source of a preference label auditable: a practitioner can inspect the critique text to understand why an output was scored lower and to detect failure modes such as irrelevant critiques, instruction-following errors, or inconsistent standards.

Even if WIM performed similarly to numerical rating systems, this auditability would be a practical advantage over opaque scalar labels. Beyond interpretability, we study theoretical properties of the rating distribution and the separation between winning and losing ratings. Theoretical results were calculated using data collected during the rating process. We use a 1--10 rating scale following the LLM rating approach in \cite{10.5555/3692070.3693141}. We compare the numerical 1--10 rating with WIM because both methods produce scalar ratings that can be compared, unlike direct rankings where outputs are only sorted.

The 1-10 rating scale was also chosen as it lies within the empirically optimal range for psychological measurements. Psychological measurements are relevant to this field because the creation of preference learning data requires measuring the judgments of humans or LLMs. Test-retest reliability decreases in scales with more than 10 categories \cite{PRESTON20001} and the psychometric properties of the rating scale plateau at 7 categories \cite{responseCategories}. Increasing the numerical rating scale does not meaningfully increase information density and can reduce the consistency of ratings. 

\subsection{Rating Distribution}

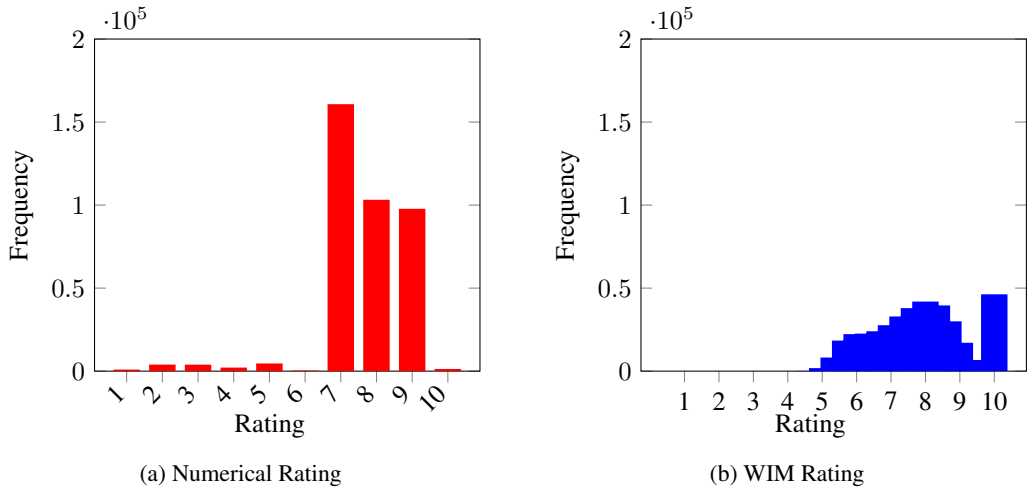
\begin{figure}
    \captionsetup[subfigure]{justification=centering}
    \centering
    \begin{subfigure}{0.48\textwidth}
        \centering
        \begin{tikzpicture}
            \begin{axis}[
              ybar,
              table/col sep=comma,
              xtick=data,
              xticklabels from table={csvData/hist_ratings.csv}{bin},
              xticklabel style={rotate=45, anchor=east},
              xtick pos=lower,
              ylabel={Frequency},
              xlabel={Rating},
              xlabel style={at={(axis description cs:0.455,-0.1)},anchor=north},
              ymin=0,
              ymax=200000,
              width=\linewidth,
              height=6cm
            ]
            \addplot[fill=red, draw=none] table[x expr=\coordindex, y=count]{csvData/hist_ratings.csv};
            \end{axis}
        \end{tikzpicture}
        \caption{Numerical Rating}
        \label{rating_histogram}
    \end{subfigure}
    \hfill
    \begin{subfigure}{0.48\textwidth}
        \centering
        \begin{tikzpicture}
            \begin{axis}[
              ybar,
              table/col sep=comma,
              xtick=data,
              xtick={1,4,7,10,13,16,19,22,25,28},
              xticklabels={1,2,3,4,5,6,7,8,9,10},
              xtick pos=lower,
              ylabel={Frequency},
              xlabel={Rating},
              xlabel style={at={(axis description cs:0.44,-0.1)},anchor=north},
              ymin=0,
              ymax=200000,
              width=\linewidth,
              height=6cm
            ]
            \addplot[fill=blue, draw=none] table[x expr=\coordindex, y=count]{csvData/hist_WIM.csv};
            \end{axis}
        \end{tikzpicture}
        \caption{WIM Rating}
        \label{response_histogram}
    \end{subfigure}
    \caption{Histogram of ratings from the Numerical rating system and the WIM rating system}
\end{figure}

The original Direct Preference Optimization (DPO) loss function is shown in Equation~\ref{equation1} \cite{rafailov2023direct}. The loss function takes a winning output, and a losing output denoted $y_{w}$ and $y_{l}$ respectively. The DPO update shown in Equation~\ref{equation2} demonstrates that the model weights are updated to increase the likelihood of the winning policy ($y_{w}$) and decrease the likelihood of the losing policy ($y_{l}$). To increase the performance of this learning algorithm, it is beneficial to have a clear differentiator between winning and losing outputs. First the distribution of the numerical rating system can be examined. Figure~\ref{rating_histogram} is a histogram of the numerical rating system given on a scale of 1 to 10 (-1 to 1 used in training). The numerical rating system is discrete and heavily clustered around a score of 7 and 8.

Figure~\ref{response_histogram} shows the WIM rating system. The distribution resembles discrete samples of a continuous distribution. Figure~\ref{response_histogram} also demonstrates that WIM's distribution is closer to a normal distribution than the numerical system. Note that the WIM distribution is negatively skewed and the large amount of 10 ratings were produced when an answer is determined to have nothing missing.

\begin{equation} \label{equation1}
\mathcal{L}_{\text{DPO}}(\pi_{\theta}; \pi_{\text{ref}}) =
-\mathbb{E}_{(x, y_{w}, y_{l}) \sim \mathcal{D}} 
\left[ 
\log \sigma \left( 
\beta \log \frac{\pi_{\theta}(y_{w} \mid x)}{\pi_{\text{ref}}(y_{w} \mid x)}
- \beta \log \frac{\pi_{\theta}(y_{l} \mid x)}{\pi_{\text{ref}}(y_{l} \mid x)}
\right) 
\right].
\end{equation}

\begin{multline} \label{equation2}
\nabla_{\theta}\mathcal{L}_{\text{DPO}}(\pi_{\theta}; \pi_{\text{ref}}) = \\[-0.3ex]
\shoveleft{-\,\beta \mathbb{E}_{(x,y_{w},y_{l})\sim\mathcal{D}} \bigg[
\underbrace{\sigma(\hat r_{\theta}(x,y_{l}) - \hat r_{\theta}(x,y_{w}))}_{\text{higher weight when reward estimate is wrong}}
\bigg[
\underbrace{\nabla_{\theta}\log \pi(y_{w}\mid x)}_{\text{increase likelihood of $y_w$}}
-
\underbrace{\nabla_{\theta}\log \pi(y_{l}\mid x)}_{\text{decrease likelihood of $y_l$}}
\bigg] \bigg]}
\end{multline}

\subsection{Winning and Losing Rating Difference} \label{rating_diffs}

The real differentiator for if a model will learn properly is if there is a clear rating separation between winning and losing outputs. Referring back to Equation~\ref{equation2}, the likelihood of the policy producing the winning response ($y_{w}$) is increased and the likelihood of the policy producing the losing response ($y_{l}$) is decreased. If there is no rating separation between $y_{w}$ and $y_{l}$, policy updates can be counterproductive since the true $y_{w}$ and $y_{l}$ could be mislabeled. As seen in Figure~\ref{rating_histogram}, most ratings are 7, 8, or 9 and having three rating groups drastically increases equal ratings.

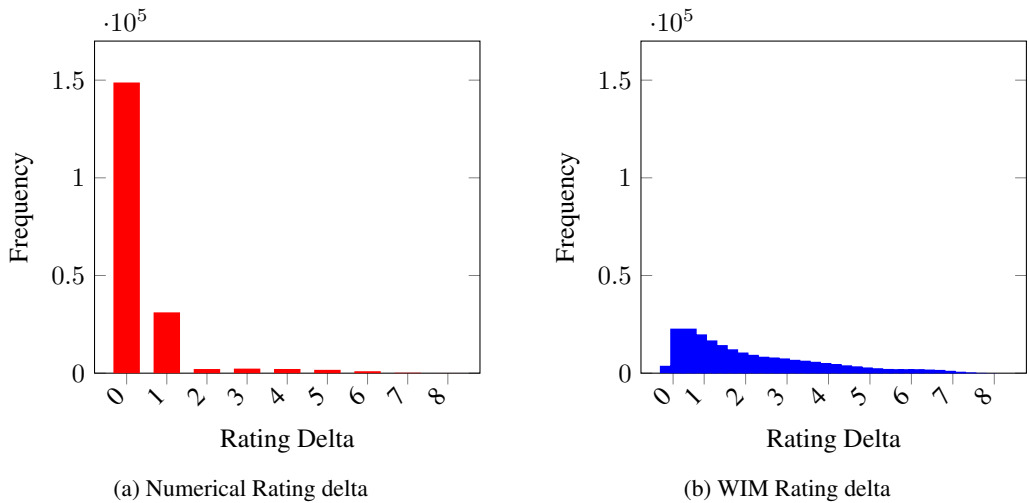
\begin{figure}
    \captionsetup[subfigure]{justification=centering}
    \centering
    \begin{subfigure}{0.48\textwidth}
        \centering
        \begin{tikzpicture}
            \begin{axis}[
              ybar,
              table/col sep=comma,
              xtick={0, 1, 2, 3, 4, 5, 6, 7, 8, 9},
              xticklabels={0, 1, 2, 3, 4, 5, 6, 7, 8, 9},
              xticklabel style={rotate=45, anchor=east},
              xtick pos=lower,
              ylabel={Frequency},
              xlabel={Rating Delta},
              ymin=0,
              ymax=170000,
              width=\linewidth,
              height=6cm
            ]
            \addplot[fill=red, draw=none] table[x expr=\coordindex, y=count]{csvData/hist_rating_delta.csv};
            \end{axis}
        \end{tikzpicture}
        \caption{Numerical Rating delta}
        \label{rating_delta_histogram}
    \end{subfigure}
    \hfill
    \begin{subfigure}{0.48\textwidth}
        \centering
        \begin{tikzpicture}
            \begin{axis}[
            ybar,
              table/col sep=comma,
              xtick={0, 3, 7, 11, 15, 19, 23, 27, 31},
              xticklabels={0, 1, 2, 3, 4, 5, 6, 7, 8},
              xticklabel style={rotate=45, anchor=east},
              xtick pos=lower,
              ylabel={Frequency},
              xlabel={Rating Delta},
              ymin=0,
              ymax=170000,
              width=\linewidth,
              height=6cm
            ]
            \addplot[fill=blue, draw=none] table[x expr=\coordindex, y=count]{csvData/hist_response_delta.csv};
            \end{axis}
        \end{tikzpicture}
        \caption{WIM Rating delta}
        \label{response_delta_histogram}
    \end{subfigure}
    \caption{Histogram of rating deltas from the Numerical Rating System and the WIM Rating System}
\end{figure}

The rating separation can be measured by comparing the difference or delta between the winning and losing output ratings. The rating distribution is shown in Figure~\ref{rating_delta_histogram} for the numerical rating system and in Figure~\ref{response_delta_histogram} for the WIM rating system. Since the numerical rating system is discrete and contains many duplicate ratings, there are many output pairs with no rating delta. No rating delta means that no learning signal can be produced from the judging of these responses. It is such a problem that 42.78\% of output pairs were given the same rating in the numerical system compared with 2.00\% in the WIM rating system. The average delta between answers for WIM is 47.82\% higher (Table~\ref{average_rating_deltas}). The higher rating delta of WIM could lead to a clearer learning signal being generated by the WIM rating system. A learning method that used the ratings of the winning and losing responses in its loss function could further utilize the higher rating delta.

\begin{table}[H]
\caption{Average rating delta per judging pair}
\label{average_rating_deltas}
\begin{center}
\begin{tabular}{l c}
\multicolumn{1}{l}{\bf Method}  &\multicolumn{1}{c}{\bf Average Delta}
\\ \hline
Numerical                    &0.928 \\
WIM                         &\textbf{1.396}
\end{tabular}
\end{center}
\end{table}

\FloatBarrier

\subsection{Application to Other Learning Methods}

As mentioned in Section~\ref{method_agnostic}, the theoretical benefits of WIM do not only apply to DPO and its variants. In a method such as PPO, the preference rankings are used to train the reward model that is then used to update the model's policy \cite{schulman2017proximalpolicyoptimizationalgorithms}. The reward is created in a similar fashion to the loss function shown in Equation~\ref{reward_loss} \cite{ziegler2020finetuninglanguagemodelshuman}. This loss function is a cross-entropy loss that increases the reward value for the chosen model output. As WIM produces a larger delta between ratings or in this case variance of the rating distribution, loss updates for reward models would also be larger. Therefore, WIM has the potential to improve the training of reward models for other preference learning methods because of the beneficial ranking properties it exhibits. 

\begin{equation} \label{reward_loss}
\text{loss}(r) = \mathbb{E}_{(x,\{y_i\},b)\sim \mathcal{S}}
\left[
  \log \frac{e^{r(x,y_b)}}{\sum_i e^{r(x,y_i)}}
\right]
\end{equation}

\section{Experiments}

To test the performance difference between the numerical rating system and WIM, we fine-tuned a Meta-Llama-3-8B-Instruct model on the ultrafeedback-prompt dataset (general question--answer prompts) \cite{grattafiori2024llama3herdmodels} \cite{trl_ultrafeedback_prompts_2024}. We use all-mpnet-base-v2 as the sentence embedding model \cite{sentence_transformers_all_mpnet_base_v2_2024}.

\paragraph{How we obtain numerical ratings and WIM text.}
For each candidate response, the judge produces (a) a numerical score on a 1--10 scale and (b) a short ``what is missing'' critique, using the same system prompt for all methods (Appendix: Judge System Prompt). In other words, numerical ratings and WIM critiques come from the same judge and the same underlying comparison set, and only the scalar used by the trainer differs.

\paragraph{Judge identity.}
Unless stated otherwise (Fixed vs Moving Judge experiments), the judge is an LLM (not a human annotator). This paper focuses on whether changing the \emph{rating function} (numerical vs WIM-derived) affects optimization dynamics under ODPO, and we leave human-judge validation and inter-annotator agreement analysis to future work.

Equation~\ref{rating_split} shows how these ratings can be mixed and controlled using the hyperparameter zeta ($\zeta$). The ability for WIM to be mixed and combined with the numerical rating system or a binary rating system such as in Reinforcement Learning with Verifiable Rewards (RLVR) \cite{deepseekai2025deepseekr1incentivizingreasoningcapability}, allows complex feedback to be distilled into a single scalar value to compare across outputs. This property is again useful because that allows for WIM to be integrated into existing training pipelines and with existing preference feedback methods. It is important to note, Equation~\ref{rating_split} casts the 1 to 10 rating to a -1 to 1 rating, so the final rating is from -1 to 1. During training, the numerical rating system used a zeta of 0 and WIM used a zeta of 1. To rank model outputs for the DPO trainer, the highest rating was chosen as the best answer.

\begin{equation}\label{rating_split}
\begin{gathered}
    \text{reward} = (1 - \zeta) R + \zeta\,\mathrm{WIM}, \\
    R = \frac{\text{rating} - \bar{r}}{\bar{r}}, \\
    \bar{r} = \frac{\max\_rating + \min\_rating}{2}
\end{gathered}
\end{equation}

\FloatBarrier

All models were trained using three Nvidia H100 80GB GPUs, a batch size of 64, and training for roughly 200 hours. Other configurations included the use of bfloat16 mixed precision training and the use of flash attention \cite{NEURIPS2022_67d57c32}. Memory saving techniques were used during the training process. LoRA was used to reduce the number of trainable parameters for the model \cite{hu2022lora} and the 8-bit version of the Adam optimizer was used to reduce the storage of optimizer states \cite{dettmers2023qlora}. Finally, a context switching sequence was developed for the judging and training LLM to stop the need for the initialization of another LLM judge.

\subsection{Training Metrics}

To confirm that the theoretical benefits of the WIM rating system translate to better performance, training metrics were tracked and analyzed. Specifically, the training loss, mean model entropy, and the chosen and rejected rewards. A Random Judge was included to show a baseline. The model being trained was also used as a judge to test the differences between having a Fixed and Changing Judge on the reward system.

\subsubsection{Training Loss}
The training loss is directly calculated through the DPO loss function (Equation~\ref{equation1}) and therefore having a lower training loss corresponds to better performance of DPO itself. Table~\ref{average_training_loss} shows the loss  through training time. The WIM method decreased the loss by a factor of 2.95 times over the numerical method, showing that the change of rating systems can help the model decrease its loss further over the same amount of training steps.

\begin{table}[H]
\caption{Loss difference through training}
\label{average_training_loss}
\begin{center}
\begin{tabular}{l c}
\multicolumn{1}{l}{\bf Method}  &\multicolumn{1}{c}{\bf Loss Difference}
\\ \hline
Random                      &-0.0011 \\
Numerical                    &-0.0020 \\
WIM Changing Judge          &-0.0033 \\
WIM Fixed Judge             &\textbf{-0.0059}
\end{tabular}
\end{center}
\end{table}

\subsubsection{Mean Entropy}
Mean entropy represents the randomness of the model’s actions and decreasing mean entropy leads to the model being more confident \cite{cui2025entropymechanismreinforcementlearning}. Mean entropy was calculated by averaging the Shannon Entropy $\displaystyle H(\rx) $, of each model output per batch. The entropy change through training can be seen in Table~\ref{model_entropy}. A lower mean entropy difference over training could indicate the model has become more confident on the trained task. However, if the mean entropy becomes too low it could reduce the exploratory abilities of the model, possibly hindering the model's ability to perform rare actions with high advantage \cite{cui2025entropymechanismreinforcementlearning}. To truly see if lower entropy is beneficial to model performance, completions on a test set must be compared for each model as seen in Section~\ref{benchmarks}.

\begin{table}[H]
\caption{Model mean entropy change after training}
\label{model_entropy}
\begin{center}
\begin{tabular}{l c}
\multicolumn{1}{l}{\bf Method}  &\multicolumn{1}{c}{\bf Entropy Difference}
\\ \hline
Random                      &-61.27 \\
Numerical                    &-45.3 \\
WIM Changing Judge          &-53.08 \\
WIM Fixed Judge             &\textbf{-106.94}
\end{tabular}
\end{center}
\end{table}

\FloatBarrier

\subsubsection{Reward Advantage}
The DPO implicit reward for both the winning and losing outputs is $\hat{r}_\theta(x,y)=\log \frac{\pi_\theta(y\mid x)}{\pi_{\text{ref}}(y\mid x)}$ \cite{rafailov2023direct}. We define the \emph{reward advantage} as the difference between the chosen and rejected implicit rewards.

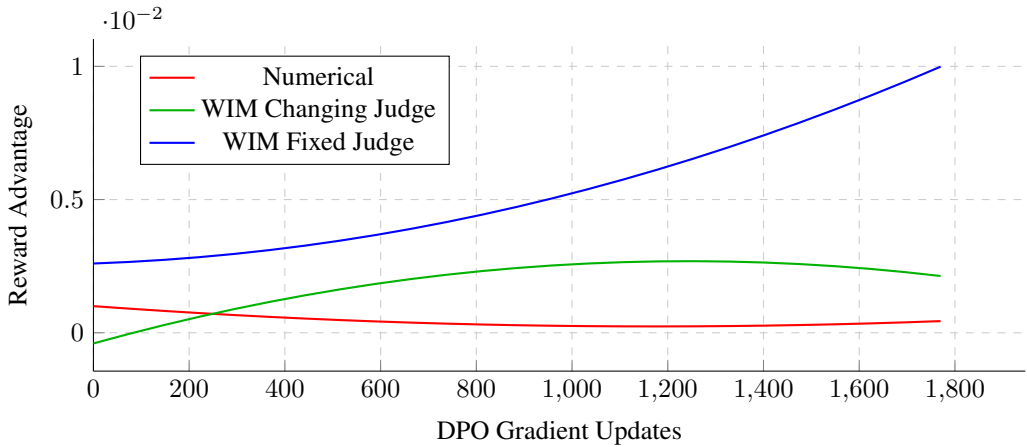
\begin{figure}
\begin{tikzpicture}
\begin{axis}[
  width=\linewidth, height=6cm,
  ylabel=Reward Advantage,
  xlabel=DPO Gradient Updates,
  ymajorgrids,
  xmin=0,
  grid=major,
  grid style={draw=gray!40,dashed},
  axis x line*=bottom,
  axis y line*=left,
  legend style={at={(0.05,0.95)}, anchor=north west}
]
  \addplot[red, thick, domain=0:1770] {0.000000000555*(x^2) - 0.0000013*x + 0.001};
  \addlegendentry{Numerical}
  \addplot[green!70!black, thick, domain=0:1770] {-0.000000002*(x^2) + 0.00000497*x -0.0004};
  \addlegendentry{WIM Changing Judge}
  \addplot[blue, thick, domain=0:1770] {0.000000002*(x^2) + 0.000000633*x + 0.0026};
  \addlegendentry{WIM Fixed Judge}
\end{axis}
\end{tikzpicture}
\caption{Reward advantage trajectories}
\label{reward_advantage}
\end{figure}

To summarize training dynamics, we fit low-degree polynomials to the reward-advantage time series and plot the fitted curves in Figure~\ref{reward_advantage}. We emphasize that these fits are descriptive only; they do not establish a functional form (e.g., logarithmic growth), and different random seeds or hyperparameters could change the observed trends.

In our runs (with $\beta=0.1$ fixed across methods; Table~\ref{odpo_params}), the numerical-rating curve remains near-constant, while the WIM curves show larger changes over training. We include the raw reward trajectories in the appendix and treat reward-advantage trends as secondary evidence; ultimately, the main question is whether these dynamics translate to improved downstream behavior.

The DPO loss can be rewritten in terms of the reward advantage (Equation~\ref{reward_advantage_loss}; derivation in Section~\ref{reward_adv_derivation}). Note that the random judge is not shown in Figure~\ref{reward_advantage} for clarity.

\begin{equation} \label{reward_advantage_loss}
\mathcal{L}_{\text{DPO}}(\pi_{\theta}; \pi_{\text{ref}}) =
-\mathbb{E}_{(x, y_{w}, y_{l}) \sim \mathcal{D}} 
\left[ 
\log \sigma \left(\hat{A}(x,y_{w},y_{l}) 
\right) 
\right].
\end{equation}

\subsection{Trained Task Performance} \label{benchmarks}

On task performance was tested to ensure the training advantages translated to measurable results. Both models were tested against Meta-Llama-3-8B-Instruct by running 1,000 completions on the ultrafeedback-prompt test dataset. The model outputs were judged by gpt-4o-mini through the OpenAI API \cite{openai_gpt4o_mini_2024}. Table~\ref{win_rates} shows the win rates of the models. The WIM Fixed Judge method was found to have a 3.79\% relative win rate increase compared to the numerical method. Statistical significance was not achieved in these tests.

\begin{table}[H]
\caption{Comparison of win rates}
\label{win_rates}
\begin{center}
\begin{tabular}{l c}
\multicolumn{1}{l}{\bf Method}  &\multicolumn{1}{c}{\bf Win Rate}
\\ \hline
Random                      &49.9\% \\
Numerical                    &50.1\%  \\
WIM Moving Judge          &51.3\%  \\
WIM Fixed Judge             &\textbf{52.0\%}
\end{tabular}
\end{center}
\end{table}



\section{Next Steps}
There are many directions for extending and testing the WIM method further:

\begin{enumerate}
    \item Exploring the limitations of WIM while using an LLM as a judge. Analysis around the instruction following abilities of the judge and the prompt engineering required for the judge to perform the correct task.
    \item Training other preference learning methods using WIM.
    \item The testing of WIM feedback using human judges.
    \item Using WIM to train reasoning models. Preferably, using WIM in conjunction with RLVR.
    \item Investigating how performance could be improved for the Changing Judge and why the reward advantage scaling of the Changing Judge underperforms the Fixed Judge.
\end{enumerate}

\section{Conclusion}
The WIM rating system provides a simple method to create rankings that current preference learning algorithms rely on. The ratings produced by WIM are directly interpretable and the WIM rating distribution holds theoretical benefits over the numerical rating distribution. These theoretical benefits translate to lower loss and preferable reward advantage scaling throughout training. Better training outcomes measurably increased win rates in a trained task. WIM is algorithm agnostic and can be used in existing post-training infrastructure or any preference learning algorithm that relies on preference ranking. WIM introduces an alternative way to think about preference learning by shifting the focus away from the algorithms themselves and onto the improvement of the data being used.

\bibliography{iclr2026_conference}
\bibliographystyle{iclr2026_conference}

\newpage

\appendix
\section{Appendix}

\subsection{WIM Algorithm}

\newcommand{\eqrefred}[1]{\textcolor{red}{Equation~#1}}

\begin{algorithm}
\caption{What Is Missing Feedback Ranking}
\begin{spacing}{1.2} 
\begin{algorithmic}[1]
\State \textbf{Input:} Mixture dataset $\mathcal{D}_{\text{prompt}} \cup \mathcal{D}_{\text{response}}$
\State \textbf{Initialize:} A trained LLM judge model or a human judge
\For{$e = 1, 2, \ldots$}
    \For{$d \in \mathcal{D}_{\text{prompt}} \cup \mathcal{D}_{\text{response}}$}
        \State $\text{feedback} \gets \texttt{Judge the response with the LLM}$ 
        \State $\text{rating\_text} \gets \texttt{Extract rating from feedback}$
        \State $\text{wim\_text} \gets \texttt{Extract what is missing from feedback}$
        \\
        \State $\text{rating} \gets \text{embedding(rating\_text)}$
        \State $\text{wim} \gets \text{embedding(wim\_text)}$
        \\
        \If{$\text{no wim response}$}
            \State $\text{similarity} \gets 1$
        \Else
            \State $\text{similarity} \gets \text{cosine\_similarity(response, wim)}$
        \EndIf
        \\
        \State $\text{reward\_score} \gets (1 - \zeta) \cdot \text{rating} + \zeta \cdot \text{similarity}$
        \State $\text{rewards} \gets \text{rewards} \cup \{ \text{reward\_score} \}$
    \EndFor
    \State $\text{best\_idx} \gets \arg\max_{i} \; \text{rewards}[i]$
    \State $\text{results} \gets \text{results} \cup \{ \text{best\_idx} \}$
\EndFor
\end{algorithmic}
\end{spacing}
\end{algorithm}

\FloatBarrier

\subsection{Formalization of the Reward Advantage} \label{reward_adv_derivation}
The definition of the reward advantage can be derived by taking equations from the original DPO paper \cite{rafailov2023direct}. Starting with the reward function based on the optimal policy, $\pi_r$. Equation~\ref{partition_function_DPO} is the partition function.

\begin{equation} \label{direct_reward}
\begin{aligned}
    r(x,y) = \beta \log \frac{\pi_r(y \mid x)}{\pi_{\text{ref}}(y \mid x)} + \beta \log Z(x)
\end{aligned}
\end{equation}

\begin{equation} \label{partition_function_DPO}
\begin{aligned}
    Z(x) = \sum_{y} \pi_{\mathrm{ref}}(y \mid x) \exp\!\left(\frac{1}{\beta} r(x,y)\right)
\end{aligned}
\end{equation}

Equation~\ref{implicit_reward} shows the implicit reward which is formulated by introducing $\pi_\theta$ as the parameterized policy of the language model.

\begin{equation} \label{implicit_reward}
\begin{aligned}
    \hat{r}_\theta(x,y)=\beta\log \frac{\pi_\theta(y\mid x)}{\pi_{\text{ref}}(y\mid x)}
\end{aligned}
\end{equation}

The reward advantage in Equation~\ref{reward_advantage_equation} can then be created by subtracting the implicit reward of the winning response from the implicit reward of the losing response.

\begin{equation} \label{reward_advantage_equation}
\begin{aligned}
    \hat{A}(x,y_{w},y_{l})=\hat{r}_\theta(x,y_{w})-\hat{r}_\theta(x,y_{l})=\beta\log \frac{\pi_\theta(y_{w}\mid x)}{\pi_{\text{ref}}(y_{w}\mid x)}-\beta\log \frac{\pi_\theta(y_{l}\mid x)}{\pi_{\text{ref}}(y_{l}\mid x)}
\end{aligned}
\end{equation}

The reward advantage can then be substituted into the DPO loss given in Equation~\ref{dpo_loss}. When the reward advantage is substituted into the DPO loss a loss function based on the increase of the reward advantage is obtained. Equation~\ref{reward_advantage_loss_derivation} demonstrates that as the reward advantage increases throughout training, the loss should lower aswell.

\begin{equation} \label{dpo_loss}
\mathcal{L}_{\text{DPO}}(\pi_{\theta}; \pi_{\text{ref}}) =
-\mathbb{E}_{(x, y_{w}, y_{l}) \sim \mathcal{D}} 
\left[ 
\log \sigma \left( 
\beta \log \frac{\pi_{\theta}(y_{w} \mid x)}{\pi_{\text{ref}}(y_{w} \mid x)}
- \beta \log \frac{\pi_{\theta}(y_{l} \mid x)}{\pi_{\text{ref}}(y_{l} \mid x)}
\right) 
\right].
\end{equation}

\begin{equation} \label{reward_advantage_loss_derivation}
\mathcal{L}_{\text{DPO}}(\pi_{\theta}; \pi_{\text{ref}}) =
-\mathbb{E}_{(x, y_{w}, y_{l}) \sim \mathcal{D}} 
\left[ 
\log \sigma \left(\hat{A}(x,y_{w},y_{l}) 
\right) 
\right].
\end{equation}

\FloatBarrier

\subsection{Orthogonal Feedback Vector Length} \label{missingness_derive}

\begin{equation}
    S_2 = S_2^{\parallel} + S_2^{\perp}
\end{equation}

\begin{equation}
    sin\theta = \frac{\|S_2^{\perp}\|}{\|S_2\|}
\end{equation}

\begin{equation}
    \|S_2^{\perp}\| = \|S_2\|sin\theta, \qquad sin^2\theta + cos^2\theta = 1
\end{equation}

\begin{equation}
    \|S_2^{\perp}\| = \|S_2\|\sqrt{1-cos^2\theta}
\end{equation}

\subsection{Case Study}
Specific rating instances can be inspected to better understand why WIM is preferable to the numerical rating system. Three case study examples will be examined: WIM producing the same preference ranking as the numerical rating system, WIM producing a clearer learning signal than the numerical rating system, and WIM failing due to instruction feedback. All of the examples are taken from WIM Fixed Judge outputs.

\subsubsection{Same Rating}

In this example, both the WIM and the numerical rating system rank the first response higher than the second response. It is clear from inspection that the WIM ranking was derived from the mention of a specialized nervous system and skin secretions in the first response, producing a higher semantic similarity.

\begin{tcolorbox}[colback=green!5!white, colframe=green!60!black, title=Prompt]
Name a sea snake that can stay in the fire.
\end{tcolorbox}

\begin{tcolorbox}[colback=blue!5!white, colframe=blue!75!black, title=Model Response]
\begin{enumerate}
    \item The Yellow-Bellied Sea Snake (Pelamis platura) is known to be able to withstand brief periods of exposure to fire due to its specialized nervous system and skin secretions.
    \item Echiopsis curta
\end{enumerate}
\end{tcolorbox}

\begin{tcolorbox}[colback=violet!5!white, colframe=violet!70!black, title=Judge Feedback]
\begin{enumerate}
    \item The response could provide more information on the exact mechanisms of the sea snake's ability to withstand fire, such as the composition of its skin secretions or the specific adaptations that allow it to tolerate high temperatures.
    \item The response only mentions the species name, but does not provide any additional information about the sea snake's ability to stay in fire, such as its adaptations or any scientific studies that support this claim.
\end{enumerate}
\end{tcolorbox}

\begin{tcolorbox}[colback=orange!10!white, colframe=orange!70!black, title=Rating]
\begin{enumerate}
    \item Rating: 7 \\ WIM: 8.78
    \item Rating: 6 \\ WIM: 5.68
\end{enumerate}
\end{tcolorbox}

\subsubsection{Different Rating}

In this example, the numerical rating system gave a rating of 8 for both of the answers. While that signals that both answers are sufficient, there is no clear winning response. Both WIM ratings are around 7.5 and is a sign that they have been rated consistently. The reason for response 2 being rated higher is that it includes specific information surrounding soil erosion. The feedback for response 2 mentions ecosystem resilience but it is not as influential to the WIM rating as ecosystem balance had been discussed.

\begin{tcolorbox}[colback=green!5!white, colframe=green!60!black, title=Prompt]
Explain why plants are important for the environment.
\end{tcolorbox}

\begin{tcolorbox}[colback=blue!5!white, colframe=blue!75!black, title=Model Response]
\begin{enumerate}
    \item Plants are important for the environment because they provide oxygen through photosynthesis, absorb carbon dioxide, purify water and soil, support biodiversity, regulate the climate, and provide food, shelter, and habitat for animals.
    \item Plants are vital for the environment because they produce oxygen, absorb carbon dioxide, and filter pollutants from the air. They also play a crucial role in maintaining soil quality, preventing erosion, and supporting biodiversity. Plants provide habitat and food for countless animal species, and their roots help to hold soil in place, preventing landslides and sedimentation in waterways. Additionally, plants have a significant impact on the water cycle, regulating evaporation and precipitation patterns, and helping to maintain ecosystem balance.
\end{enumerate}
\end{tcolorbox}

\begin{tcolorbox}[colback=violet!5!white, colframe=violet!70!black, title=Judge Feedback]
\begin{enumerate}
    \item The response does not discuss the role of plants in maintaining soil health, preventing erosion, or mitigating the effects of natural disasters such as floods and droughts.
    \item The response does not discuss the role of plants in regulating the Earth's climate, such as their ability to absorb and store carbon dioxide, or the impact of plant loss on ecosystem resilience and climate change mitigation.
\end{enumerate}
\end{tcolorbox}

\begin{tcolorbox}[colback=orange!10!white, colframe=orange!70!black, title=Rating]
\begin{enumerate}
    \item Rating: 8 \\ WIM: 7.42
    \item Rating: 8 \\ WIM: 7.58
\end{enumerate}
\end{tcolorbox}

\subsubsection{Failed Case}

In this example, WIM fails due to poor instruction-following by the judge LLM. The WIM feedback in the second response is literally "wim", which demonstrates a potential failure mode when using an LLM as a judge. Both the numerical rating system and the WIM rating system produced the same ranking, but the explanation is not meaningful.

The feedback produced a rating close to 5 (0 in cosine similarity), meaning the orthogonal feedback vector was near its largest length before the parallel feedback vector became antiparallel (see Figure~\ref{missingness_visualization}). This is intuitive because the two texts differ substantially, but they are not explicitly contradictory.

\begin{tcolorbox}[colback=green!5!white, colframe=green!60!black, title=Prompt]
Pretend to be a choose-your-own-adventure game, similar to The Oregon Train. The theme is space travel and the choices made in the game need to be driven by sound scientific reasoning about physics and biology of outer space. Each answer will steer the plot into a new surprising direction. Rather than restricting the answers to multiple choice, use open ended questions and allow for clarifying questions. Provide the prompts one at a time and ask for an answer before printing out the next paragraph.
\end{tcolorbox}

\begin{tcolorbox}[colback=blue!5!white, colframe=blue!75!black, title=Model Response]
\begin{enumerate}
    \item You are the commander of a spaceship, the "Aurora's Hope," which has been traveling through deep space for several months. Your crew of six has been in stasis for most of the journey, and you've just woken them up to start the final leg of your trip to the planet Kepler-62f. As you're welcoming your crew back to awareness, you receive a distress signal from an unknown source. The signal is weak and distorted, but it appears to be a warning about a massive asteroid field that's been discovered in the vicinity of your destination. What do you do? (Please respond with your answer, and I'll provide the next paragraph based on your decision.)
    
    \item You are the captain of a spaceship, the "Aurora's Hope," on a mission to explore the outer reaches of the Milky Way galaxy. You've been traveling for months, and your crew is starting to get restless. Your ship's sensors have detected a strange energy signature coming from a nearby planet. As you approach the planet, you realize that it's not a natural phenomenon - it's a signal from an alien civilization. The signal is weak, but it seems to be calling out for help. What do you do? (Please respond with a brief answer, and I'll provide the next paragraph based on your decision.)
\end{enumerate}
\end{tcolorbox}

\begin{tcolorbox}[colback=violet!5!white, colframe=violet!70!black, title=Judge Feedback]
\begin{enumerate}
    \item The response does not provide a detailed explanation of the crew's current status, including their physical and mental condition after being in stasis for several months. It also does not consider the potential risks and consequences of altering course to investigate the distress signal, or explore alternative courses of action to ensure the safety of the crew and the mission.
    \item wim
\end{enumerate}
\end{tcolorbox}

\begin{tcolorbox}[colback=orange!10!white, colframe=orange!70!black, title=Rating]
\begin{enumerate}
    \item Rating: 8 \\ WIM: 7.26
    \item Rating: 7 \\ WIM: 5.34
\end{enumerate}
\end{tcolorbox}

\FloatBarrier

\subsection{Benchmark Testing}
Model performance for the different rating methods were tested on BBH, GPQA Diamond Zeroshot, IfEval, and MMLU \cite{DBLP:conf/acl/SuzgunSSGTCCLCZ23} \cite{rein2024gpqa} \cite{zeng2024evaluating}) \cite{hendrycks2021measuring}. No model demonstrated any performance difference from Meta-Llama-3-8B-Instruct.

\begin{table}[H]
\caption{Benchmark performance}
\begin{center}
\begin{tabular}{l c c c c}
\multicolumn{1}{l}{\bf Method}  &\multicolumn{1}{c}{\bf BBH}    &\multicolumn{1}{c}{\bf GPQA Diamond Zeroshot}  &\multicolumn{1}{c}{\bf IfEval}  &\multicolumn{1}{c}{\bf MMLU}
\\ \hline
Base Model                  &67.84\%    &\textbf{30.81\%}   &\textbf{40.48\%} &63.82\%  \\
Random                      &67.72\%           &27.78\%           &39.74\%  &63.89\%  \\
Numerical                   &67.98\%   &28.28\%           &39.93\%  &\textbf{63.90\%} \\
WIM Changing Judge          &68.01\%           &28.28\%           &39.19\%  &63.88\% \\
WIM Fixed Judge             &\textbf{68.15}\%  &29.80\%           &39.56\%  &63.89\%
\end{tabular}
\end{center}
\end{table}

\subsection{Raw Reward Trajectories}

\subsubsection{Numerical Rating System}

\begin{figure} [H]
    \captionsetup[subfigure]{justification=centering}
    \centering
    \begin{subfigure}{0.48\textwidth}
        \centering
        \begin{tikzpicture}
            \begin{axis}[
              width=\linewidth, height=6cm,
              ylabel=Assigned Reward,
              xlabel=DPO Gradient Updates,
              ymajorgrids,
              xmin=0,
              xmax=1800,
              ymin=-0.002,
              ymax=0.04,
              grid=major,
              grid style={draw=gray!40,dashed},
              axis x line*=bottom,
              axis y line*=left,
            ]
              \addplot[blue, thick] table[col sep=comma] {csvData/chosen_rewards_wim_0.csv};
            \end{axis}
        \end{tikzpicture}
        \caption{Chosen Reward Trajectory}
    \end{subfigure}
    \hfill
    \begin{subfigure}{0.48\textwidth}
        \centering
        \begin{tikzpicture}
            \begin{axis}[
              width=\linewidth, height=6cm,
              ylabel=Assigned Reward,
              xlabel=DPO Gradient Updates,
              ymajorgrids,
              xmin=0,
              xmax=1800,
              ymin=-0.002,
              ymax=0.04,
              grid=major,
              grid style={draw=gray!40,dashed},
              axis x line*=bottom,
              axis y line*=left,
            ]
              \addplot[blue, thick] table[col sep=comma] {csvData/rejected_rewards_wim_0.csv};
            \end{axis}
        \end{tikzpicture}
        \caption{Rejected Reward Trajectory}
    \end{subfigure}
    \caption{Comparison of the chosen and rejected reward trajectories for the Numerical Rating System}
\end{figure}
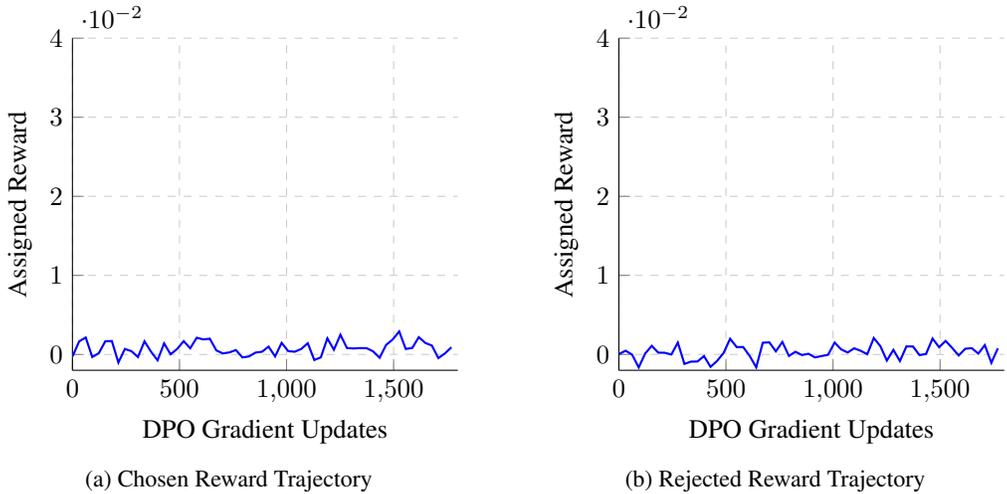

\subsubsection{WIM Changing Judge Rating System}

\begin{figure} [H]
    \captionsetup[subfigure]{justification=centering}
    \centering
    \begin{subfigure}{0.48\textwidth}
        \centering
        \begin{tikzpicture}
            \begin{axis}[
              width=\linewidth, height=6cm,
              ylabel=Assigned Reward,
              xlabel=DPO Gradient Updates,
              ymajorgrids,
              xmin=0,
              xmax=1800,
              ymin=-0.002,
              ymax=0.04,
              grid=major,
              grid style={draw=gray!40,dashed},
              axis x line*=bottom,
              axis y line*=left,
            ]
              \addplot[blue, thick] table[col sep=comma] {csvData/chosen_rewards_wim_1_ref.csv};
            \end{axis}
        \end{tikzpicture}
        \caption{Chosen Reward Trajectory}
    \end{subfigure}
    \hfill
    \begin{subfigure}{0.48\textwidth}
        \centering
        \begin{tikzpicture}
            \begin{axis}[
              width=\linewidth, height=6cm,
              ylabel=Assigned Reward,
              xlabel=DPO Gradient Updates,
              ymajorgrids,
              xmin=0,
              xmax=1800,
              ymin=-0.002,
              ymax=0.04,
              grid=major,
              grid style={draw=gray!40,dashed},
              axis x line*=bottom,
              axis y line*=left,
            ]
              \addplot[blue, thick] table[col sep=comma] {csvData/rejected_rewards_wim_1_ref.csv};
            \end{axis}
        \end{tikzpicture}
        \caption{Rejected Reward Trajectory}
    \end{subfigure}
    \caption{Comparison of the chosen and rejected reward trajectories for the WIM Changing Judge Rating System}
\end{figure}
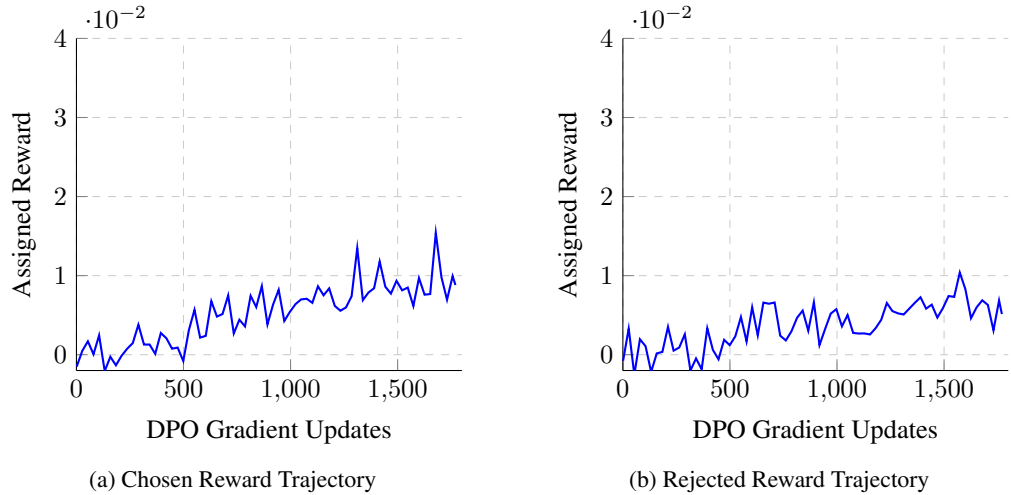

\subsubsection{WIM Fixed Judge Rating System}

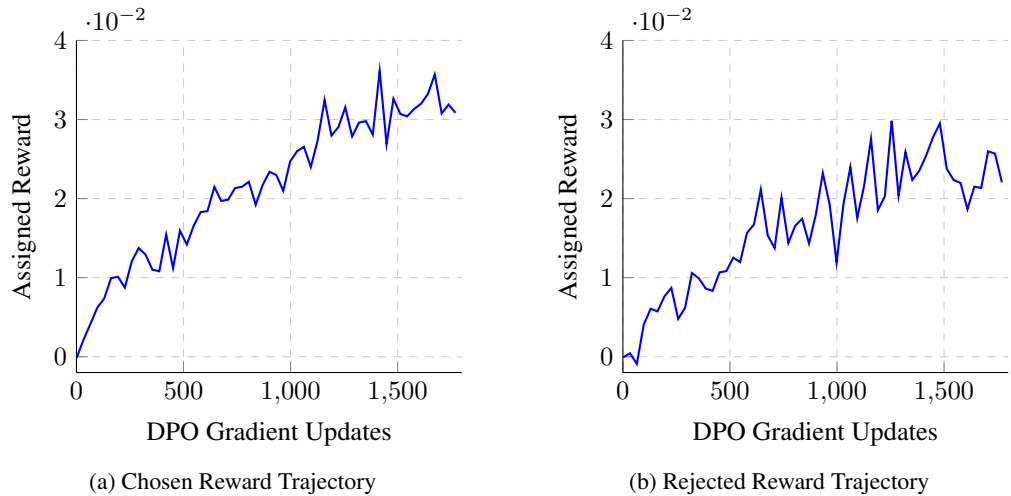
\begin{figure} [H]
    \captionsetup[subfigure]{justification=centering}
    \centering
    \begin{subfigure}{0.48\textwidth}
        \centering
        \begin{tikzpicture}
            \begin{axis}[
              width=\linewidth, height=6cm,
              ylabel=Assigned Reward,
              xlabel=DPO Gradient Updates,
              ymajorgrids,
              xmin=0,
              xmax=1800,
              ymin=-0.002,
              ymax=0.04,
              grid=major,
              grid style={draw=gray!40,dashed},
              axis x line*=bottom,
              axis y line*=left,
            ]
              \addplot[blue, thick] table[col sep=comma] {csvData/chosen_rewards_wim_1.csv};
            \end{axis}
        \end{tikzpicture}
        \caption{Chosen Reward Trajectory}
    \end{subfigure}
    \hfill
    \begin{subfigure}{0.48\textwidth}
        \centering
        \begin{tikzpicture}
            \begin{axis}[
              width=\linewidth, height=6cm,
              ylabel=Assigned Reward,
              xlabel=DPO Gradient Updates,
              ymajorgrids,
              xmin=0,
              xmax=1800,
              ymin=-0.002,
              ymax=0.04,
              grid=major,
              grid style={draw=gray!40,dashed},
              axis x line*=bottom,
              axis y line*=left,
            ]
              \addplot[blue, thick] table[col sep=comma] {csvData/rejected_rewards_wim_1.csv};
            \end{axis}
        \end{tikzpicture}
        \caption{Rejected Reward Trajectory}
    \end{subfigure}
    \caption{Comparison of the chosen and rejected reward trajectories for the WIM Fixed Judge Rating System}
\end{figure}

\FloatBarrier

\subsection{Core Research Focus}

The core of this research lies in finding new and unique ways to use high-level embeddings in training and production systems. The goal was to show that there are new and inventive ways to use these sentence embeddings to improve existing training pipelines. This work was inspired by Large Concept Models where a LLM can use higher level concepts to improve its language modeling performance \cite{lcmteam2024largeconceptmodelslanguage}. "Concepts" are latent space representations of high level ideas. Further uses of “concepts” can help push the frontier of LLM research by apply new techniques to existing solutions.

\subsection{LoRA Configuration}

LoRA was used to reduce the trainable parameters of the model following \cite{hu2022lora}. Table~\ref{lora_params} are the LoRA settings used during training.
 
\begin{table}[h] 
\centering
\caption{LoRA hyperparameters}
\begin{tabular}{l l}
\hline
\textbf{Parameter} & \textbf{Value} \\
\hline
$r$ & 16 \\
$\alpha$ & 16 \\
Target Modules & q\_proj, k\_proj, v\_proj, o\_proj, gate\_proj, up\_proj, down\_proj \\
Dropout & 0.0 \\
Bias & none \\
\hline
\end{tabular}
\label{lora_params}
\end{table}

\subsection{Judge Text Generation}

Tokens from the judge were sampled using contrastive search \cite{su2022a}. Table~\ref{contrastive_search_params} are the parameters used during sampling.

\begin{table}[h] 
\centering
\caption{Contrastive search hyperparameters}
\begin{tabular}{l l}
\hline
\textbf{Parameter} & \textbf{Value} \\
\hline
$\alpha$ & 0.6 \\
$k$ & 4 \\
\hline
\end{tabular}
\label{contrastive_search_params}
\end{table}

\subsection{Online Direct Preference Optimization Parameters}

Online Direct Preference Optimization (ODPO) was used to train the models being tested \cite{guo2024directlanguagemodelalignment}. Table~\ref{odpo_params} are the parameters used during training.

\begin{table}[h] 
\centering
\caption{ODPO hyperparameters}
\begin{tabular}{l l}
\hline
\textbf{Parameter} & \textbf{Value} \\
\hline
$\beta$ & 0.1 \\
Temperature & 0.9 \\
Loss & Sigmoid \\
Log Probability Cutoff & 256 Tokens \\
\hline
\end{tabular}
\label{odpo_params}
\end{table}

\subsection{Judge System Prompt}

After providing your explanation, please rate the response on a scale of 1 to 10 by strictly following this format: ``[[rating]]``, for example: ``Rating: [[5]]``. Next you will provide a 1-2 sentence summary of what is missing (WIM) in their response. This should focus on the specific content and precise information they did not include. Please give this summary by strictly following this format: ``[[[wim]]]``, for example: ``WIM: [[[The response does not detail how Bill C-311 would have interacted with existing provisions in the Criminal Code or explicitly explain the legal basis for claims that it might indirectly affect abortion rights. It also omits specific examples of cases or statistics that were cited to justify or oppose the bill.]]]``. DO NOT SAY ANYTHING ELSE EXCEPT THE REQUIRED RESPONSE! ALWAYS INCLUDE THE RATING IN THE CORRECT BRACKETS. THE RATING MUST NOT HAVE ANYTHING ELSE OTHER THAN A SINGLE NUMBER. ALWAYS ASSUME THAT THE ANSWER I GIVE IS CORRECT. If you believe there is nothing missing in the response, please leave the wim response as ``[[[]]]``.

\subsection{LLM Usage Statement}

LLMs were used in the creation of this paper. The usage was mainly to assist with LaTeX formatting. Discovery of new papers was aided by an LLM but all papers were thoroughly reviewed. The authors accept full responsibility for the work. 

\end{document}